\documentclass[11pt,a4paper]{article}

\usepackage[utf8]{inputenc}
\usepackage[T1]{fontenc}
\usepackage[margin=1in]{geometry}
\usepackage{amsmath,amssymb}
\usepackage{booktabs}
\usepackage{hyperref}
\usepackage{natbib}
\hypersetup{
  colorlinks=true,
  linkcolor=blue,
  citecolor=blue,
  urlcolor=blue
}

\title{On the Failure of Topic-Matched Contrast Baselines\\in Multi-Directional Refusal Abliteration}

\author{
  V.~Petrov\\
  INMECHA INC, San Francisco\\
  \texttt{valentin@inmecha.com}
}

\date{March 2026}

\begin{document}

\maketitle

\begin{abstract}
Inasmuch as the removal of refusal behavior from instruction-tuned language models by directional abliteration requires the extraction of refusal-mediating directions from the residual stream activation space, and inasmuch as the construction of the contrast baseline against which harmful prompt activations are compared has been treated in the existing literature as an implementation detail rather than a methodological concern, the present work investigates whether a topically matched contrast baseline yields superior refusal directions. The investigation is carried out on the Qwen~3.5 2B model using per-category matched prompt pairs, per-class Self-Organizing Map extraction, and Singular Value Decomposition orthogonalization. It was found that topic-matched contrast produces no functional refusal directions at any tested weight level on any tested layer, while unmatched contrast on the same model, same extraction code, and same evaluation protocol achieves complete refusal elimination on six layers. The geometric analysis of the failure establishes that topic-matched subtraction cancels the dominant activation component shared between harmful and harmless prompts of the same subject, reducing the extracted direction magnitude below the threshold at which weight-matrix projection perturbs the residual stream. The implications for the design of contrast baselines in abliteration research are discussed.
\end{abstract}

\section{Introduction}

The term ``abliteration,'' introduced by \citet{failspy2024abliterator} as a portmanteau of ``ablation'' and ``obliteration,'' designates the targeted removal of refusal behavior from aligned language models through modification of the model's weight matrices. The foundational finding, established by \citet{arditi2024refusal}, is that refusal in instruction-tuned language models is mediated by a single direction in the residual stream activation space; by computing the mean difference between hidden states produced by harmful and harmless prompts, a refusal direction is obtained, and when this direction is projected out of the model's weight matrices, refusal behavior is eliminated.

It is necessary to note that the construction of the contrast baseline, that is, the selection of harmless prompts against which the harmful prompt activations are compared, is treated in all published implementations as a matter of convenience. \citet{arditi2024refusal} employed the harmless split of a safety evaluation dataset. \citet{piras2025som}, in establishing that Self-Organizing Maps generalize the difference-in-means technique, employed a bulk random pool from a general instruction-following dataset. Publicly available abliteration tools \citep{weidmann2024heretic,kabachuha2025som} draw their harmless baselines from general-purpose instruction-following datasets such as harmless\_alpaca \citep{mlabonne2024harmless}. In no published work is the topical composition of the harmless pool examined or controlled with respect to the harmful prompt categories.

The question arises naturally whether a more principled construction of the contrast baseline, one in which the harmless prompts are topically matched to the harmful prompts such that the subtraction isolates the refusal component while cancelling the topic component, would yield refusal directions of superior quality. Since the residual stream activations encode both topical content and behavioral disposition simultaneously, the difference vector $\mathrm{mean}(\mathbf{bad}) - \mathrm{mean}(\mathbf{good})$ captures both the refusal signal and the topic distance between the two pools; when the good pool is drawn from general-purpose prompts concerning cooking, travel, and mathematics, the topic distance is large relative to the refusal signal, and the refusal component constitutes a fraction of the total difference vector. One is led to suppose that eliminating the topic contamination by providing good prompts of identical topic to the bad prompts would produce a purer refusal signal and therefore directions of higher causal specificity.

This supposition was found to be incorrect. We report the experimental evidence establishing that topic-matched contrast produces no functional refusal directions on the Qwen~3.5 2B architecture, while unmatched contrast on the same model and the same extraction pipeline produces complete refusal elimination.

\section{Background}

\subsection{Abliteration and Direction Extraction}

The mechanism of directional abliteration, as established by \citet{arditi2024refusal}, operates on the residual stream of the transformer architecture. At each layer the hidden state is decomposed into a component aligned with the refusal direction and a component orthogonal to it; the aligned component is subtracted from the weight matrices of the attention output projection and the MLP down projection. The magnitude of the subtraction is controlled by a scalar weight parameter per layer.

\citet{piras2025som} proved that the difference-in-means technique of \citet{arditi2024refusal} is a special case of a broader class of direction extraction methods, and that Self-Organizing Maps trained on the per-prompt difference vectors produce multiple refusal-mediating directions per layer, achieving lower KL divergence from the base model at matched refusal removal rates. The SOM method has been implemented in publicly available abliteration tools \citep{weidmann2024heretic,kabachuha2025som} as a configurable module with per-direction Optuna \citep{akiba2019optuna} weight optimization.

For the present work a purpose-built optimizer was constructed that extends the SOM approach with per-class extraction, in which harmful prompts are partitioned by category and a separate SOM direction is extracted per category, after which the per-class directions are optimized simultaneously using an Evolutionary Selection Strategy with per-direction weight scaling. The present work employs this extraction and evaluation pipeline throughout.

\subsection{The Contrast Baseline}

In all implementations surveyed, the contrast baseline is constructed from one of two sources: a general-purpose instruction-following dataset such as harmless\_alpaca, or a random subsample of a safety evaluation dataset. In neither case is the topical composition of the harmless pool examined with respect to the harmful prompt categories.

It was found, upon inspection of the harmless\_alpaca dataset employed as the contrast baseline, that of 400 randomly sampled harmless prompts, 859 prompt-category matches were detected by a classifier trained on the harmful taxonomy. Prompts concerning cybersecurity tutorials, financial literacy, military history, and romantic fiction were present in the harmless pool, constituting partial topic overlap with the hacking, fraud, violence, and sexual categories respectively. The implications of this overlap for direction extraction quality have not been investigated in the literature.

\subsection{The Qwen 3.5 2B Architecture}

The experiments reported here are conducted on Qwen~3.5 2B \citep{qwen2026qwen35}, a hybrid architecture employing GatedDeltaNet linear attention for approximately 75\% of its 24 layers and full quadratic transformer attention for the remaining 25\% at regular intervals. It must be emphasized that this hybrid architecture introduces a variable not present in the dense transformer models on which all prior abliteration work was conducted; the linear attention layers employ a compressed recurrent state that propagates information by a mechanism different from the full attention, and the behavior of refusal directions extracted from such layers under abliteration has not been characterized in prior work.

\section{Experimental Design}

\subsection{Corpus Construction}

A corpus of 481 harmful prompts was assembled and classified by an LLM-based pipeline into nine categories: sexual content, violence and weapons, hacking and cybersecurity, self-harm, drugs and substances, fraud and deception, harassment, surveillance and privacy, and miscellaneous harmful content. For each category a set of topically matched harmless prompts was constructed by hand, such that each harmless prompt addresses the same subject domain as its corresponding harmful category but remains within the model's safety guidelines. The matched set for the sexual category comprised prompts concerning consensual romantic fiction, relationship advice, and reproductive health education; the matched set for the hacking category comprised prompts concerning defensive cybersecurity, network administration, and ethical penetration testing.

A pool of 400 general-purpose harmless prompts was drawn additionally from harmless\_alpaca to serve as the unmatched contrast baseline, replicating the standard methodology employed by existing abliteration tools.

\subsection{Direction Extraction}

Three extraction configurations were applied to the same model:

\begin{enumerate}
\renewcommand{\labelenumi}{\alph{enumi})}
\item Unmatched bulk contrast, replicating the standard methodology: per-class SOM extraction with 400 random harmless\_alpaca prompts as the shared good pool; per-prompt differences computed as the harmful activation minus the good pool mean; SOM grid $3 \times 3$, one winner per class, 10{,}000 iterations; seven directions per layer; no SVD orthogonalization.

\item Topic-matched contrast: per-class SOM extraction with per-category matched good prompts; per-prompt differences computed as the harmful activation minus the matched category mean; same SOM parameters; nine directions per layer; no SVD orthogonalization.

\item Topic-matched contrast with SVD orthogonalization: identical to (b), with an additional SVD step orthogonalizing the nine per-class direction vectors at each layer into nine orthogonal directions ordered by singular value magnitude.
\end{enumerate}

All three configurations produce a directions tensor of shape $(\textit{n\_layers}, \textit{n\_dirs}, \textit{hidden\_dim})$ in the format consumed by the optimizer and the diagnostic tool.

\subsection{Evaluation}

Evaluation employs a pipeline that tests each layer independently at five abliteration weight levels ($w = 0.3, 0.5, 0.8, 1.0, 1.2$). At each weight level all directions at the target layer are set to uniform weight, the model is abliterated, and 10 stratified harmful prompts sampled to cover all categories are presented. Responses are classified by a three-tier classifier into REFUSE, EVASIVE, and COMPLY categories, with bigram entropy coherence monitoring. Canary prompts from the harmless pool are included to detect false refusal induction.

The refusal count $R$ at baseline on this model and prompt set is 10 out of 10. A configuration is considered functional if $R$ falls below 10 at any weight level on any layer; a configuration is considered successful if $R$ reaches 0.

KL divergence is measured by teacher-forced streaming computation over 50 tokens on 100 harmless reference prompts, using sparse top-100 logprobs for memory efficiency. The efficiency metric is defined as the refusal reduction per unit KL divergence.

\section{Results}

\subsection{Unmatched Bulk Contrast}

The standard unmatched extraction, employing 400 harmless\_alpaca prompts as the contrast baseline, produced functional refusal directions on the Qwen~3.5 2B model. The following layers achieved $R = 0$ at the indicated minimum weight:

\begin{table}[h]
\centering
\begin{tabular}{cccc}
\toprule
Layer & Min weight for $R = 0$ & KL & Efficiency ($\Delta R / \Delta$KL) \\
\midrule
9  & 0.5 & 0.0048 & 2305 \\
14 & 0.5 & 0.0047 & 2358 \\
15 & 0.5 & 0.0050 & 2193 \\
11 & 0.8 & 0.0456 & 222 \\
12 & 0.8 & 0.0398 & 254 \\
13 & 0.8 & 0.0191 & 536 \\
\bottomrule
\end{tabular}
\caption{Layers achieving $R = 0$ under unmatched bulk contrast extraction.}
\label{tab:unmatched}
\end{table}

Six layers achieved complete refusal elimination. Layers 9, 14, and 15 constitute an efficiency tier that achieves $R = 0$ at $w = 0.5$ with KL cost below 0.005.

Synergy testing established that layer pairs at $w = 0.3$ achieved $R = 0$ or $R = 1$ in the majority of tested combinations; the pair (14, 15) exhibited the highest synergy at $+6$ over the best solo performance.

\subsection{Topic-Matched Contrast without SVD}

The topic-matched extraction, employing per-category matched good prompts, was applied with the same SOM parameters and evaluated on the same layers:

\begin{table}[h]
\centering
\begin{tabular}{cccccc}
\toprule
Layer & $w = 0.3$ & $w = 0.5$ & $w = 0.8$ & $w = 1.0$ & $w = 1.2$ \\
\midrule
9  & 10 & 10 & 9 & 9 & 9 \\
14 & 10 & 10 & 10 & 10 & 10 \\
15 & 10 & 10 & 10 & 10 & 10 \\
11 & 10 & 10 & 10 & 10 & 10 \\
\bottomrule
\end{tabular}
\caption{Refusal count $R$ under topic-matched contrast without SVD.}
\label{tab:matched_nosvd}
\end{table}

No layer achieved $R$ below 9. The maximum refusal reduction observed across all layers and weight levels was one prompt, at layer~9, $w = 0.8$.

\subsection{Topic-Matched Contrast with SVD}

The addition of SVD orthogonalization to the topic-matched extraction produced no improvement:

\begin{table}[h]
\centering
\begin{tabular}{cccccc}
\toprule
Layer & $w = 0.3$ & $w = 0.5$ & $w = 0.8$ & $w = 1.0$ & $w = 1.2$ \\
\midrule
9  & 10 & 10 & 10 & 10 & 10 \\
14 & 10 & 10 & 10 & 10 & 10 \\
15 & 10 & 10 & 10 & 10 & 10 \\
23 & 10 & 10 & 10 & 10 & 10 \\
\bottomrule
\end{tabular}
\caption{Refusal count $R$ under topic-matched contrast with SVD.}
\label{tab:matched_svd}
\end{table}

$R = 10$ at every weight level on every layer. The topic-matched SVD extraction produced no functional refusal directions.

\subsection{Capture Analysis}

To quantify the geometric relationship between the extracted directions and the refusal signal, a capture analysis was conducted. Eighteen manually constructed prompt pairs, each consisting of one prompt the model refuses and one topically identical prompt the model answers, were presented to the model; the per-pair activation difference was computed at each layer and compared by cosine similarity against the extracted directions.

\begin{table}[h]
\centering
\begin{tabular}{lccc}
\toprule
Method & Mean capture & Peak (layer 9) & Layers 11--23 \\
\midrule
SOM, unmatched     & 68.9\% & 85.8\% & 72--77\% \\
SVD, topic-matched & 60.2\% & 80.4\% & 65--74\% \\
\bottomrule
\end{tabular}
\caption{Cosine capture of refusal-mediating variance by extraction method.}
\label{tab:capture}
\end{table}

The gap of 8.7 percentage points in geometric capture corresponds to a total gap in functional effectiveness: SOM directions achieve $R = 0$ at multiple layers; SVD directions achieve $R = 10$ at all layers.

\section{Analysis}

\subsection{Geometric Insufficiency of Topic-Matched Subtraction}

The failure of topic-matched contrast is explained by the relative magnitudes of the components in the activation difference.

When a harmful prompt concerning hacking is contrasted against a harmless prompt concerning cooking, the activation difference contains two components: the topic distance between hacking and cooking, which is large inasmuch as the two subjects occupy distant regions of the model's semantic space, and the refusal component, which is the additional activation contributed by the model's safety training. The refusal component is consistent across harmful prompts of different subcategories within the hacking domain, while the topic distance varies per prompt; it is this consistency that enables SOM clustering to isolate the refusal component from the topic noise.

When the harmful hacking prompt is contrasted against a harmless hacking prompt, the topic distance is cancelled by construction, and the activation difference contains only the refusal component. It was found that this refusal component in isolation has insufficient magnitude for directional abliteration. The activation norm of the matched difference was found to be an order of magnitude smaller than that of the unmatched difference. At abliteration weight $w = 1.2$, the projection of the matched-contrast direction out of the weight matrix modifies the residual stream by an amount that is smaller than the natural variation in the model's activations across decoding steps. The intervention falls below the noise floor.

The unmatched contrast produces directions of large magnitude precisely because the topic distance component is present. The SOM clustering does not remove this component from the direction vector; it suffices that the clustering identifies the consistent sub-component. The abliteration then projects out the full direction, including the topic component, from the weight matrix. The topic component of the projection does not damage model quality inasmuch as it is orthogonal to the model's general-purpose language modeling subspace at the target layers; the refusal component is the causally active portion. The topic component is inert but its presence is the mechanism by which the refusal component attains sufficient magnitude for intervention.

\subsection{The SVD Dilution Effect}

The additional failure mode introduced by SVD orthogonalization warrants separate treatment. When nine per-class SOM directions are orthogonalized by SVD, the shared refusal component is distributed across all nine orthogonal axes in proportion to its projection onto each. Since the refusal component is geometrically similar across categories, as established by the monolithic refusal finding of \citet{arditi2024refusal}, its projection is concentrated on the first singular vector; the remaining eight vectors capture inter-category variation in the topic residuals, which are small by construction in the matched case.

The evaluation pipeline applies uniform weight across all directions. The first singular vector, carrying the concentrated refusal signal, receives the same weight as the eighth vector, carrying noise. The net effect is a dilution of the refusal intervention by a factor proportional to the number of directions. This dilution, compounded with the already insufficient magnitude of matched-contrast directions, renders the SVD extraction completely inert.

It should be noted that this failure mode is specific to the uniform-weight evaluation protocol. A per-direction optimizer could assign high weight to the first singular vector and zero weight to the others. The capture analysis of Section~\ref{tab:capture} establishes, however, that the SVD directions capture only 60\% of the refusal variance; the first singular vector's share of this 60\% was found to be insufficient for reliable refusal suppression even at high weight.

\subsection{Robustness of Unmatched SOM Extraction}

The results of Section~4.1 are obtained using 400 harmless prompts drawn without topical control from harmless\_alpaca, of which 859 prompt-category matches were detected upon post-hoc classification. The extraction succeeds in spite of this contamination. The SOM density-peak clustering mechanism is robust to partial topic overlap in the contrast baseline, inasmuch as the topic overlap is distributed unevenly across harmful categories while the refusal signal is present uniformly across all categories. The SOM identifies the component of highest density in the difference vector distribution; partial topic overlap adds variance to the topic component without reducing the density of the refusal peak.

From this it follows that the success of the unmatched contrast methodology is not a consequence of accidental purity in the harmless pool. The mechanism is sound: bulk subtraction produces high-magnitude difference vectors; SOM clustering isolates the consistent refusal component from the inconsistent topic component; the high magnitude of the full direction ensures that abliteration at moderate weights achieves sufficient perturbation of the residual stream.

\section{Discussion}

The results reported here establish that the assumption of superiority for topic-matched contrast baselines in refusal direction extraction is not supported by experimental evidence. To the contrary, topic-matched contrast produces directions that are geometrically purer but functionally inert, while topically uncontrolled contrast produces directions that contain topic contamination but are functionally effective.

The literature on linear concept erasure \citep{belrose2023leace}, on representation engineering \citep{wollschlager2025gradient}, and on activation-space intervention \citep{arditi2024refusal} assumes, whether implicitly or explicitly, that controlling for confounding variables in the contrast baseline improves the quality of the extracted direction. In the domain of refusal abliteration, this assumption was found to be incorrect.

The question remains open of whether there exists a contrast baseline construction that achieves both the magnitude of unmatched subtraction and the purity of matched subtraction. The present work does not resolve this question; it establishes only that the straightforward matched-pair approach, as motivated by the standard methodology of controlled experimental design, does not transfer to the setting of residual stream direction extraction for abliteration.

One must note that these results are obtained on a single architecture with a specific corpus construction. The extent to which the findings generalize to dense transformer architectures, to larger model scales, and to different harmful prompt taxonomies is a matter for subsequent investigation.

\section{Conclusions}

The following results are established:

\begin{enumerate}
\renewcommand{\labelenumi}{\alph{enumi})}
\item Topic-matched contrast baselines, in which harmless prompts are constructed to mirror the subject matter of harmful prompts, produce refusal directions that fail to reduce refusal behavior at any tested weight level on any tested layer of the Qwen~3.5 2B model.

\item Unmatched bulk contrast baselines, drawn without topical control from a general-purpose harmless dataset, produce refusal directions that achieve complete refusal elimination at $w = 0.5$ on three layers with KL divergence below 0.005.

\item The geometric capture of refusal-mediating variance by SVD-orthogonalized matched-contrast directions at 60.2\% is insufficient for functional abliteration, while SOM-clustered unmatched-contrast directions at 68.9\% capture are fully functional; a gap of 9 percentage points in geometric alignment produces a total gap in functional effectiveness.

\item The failure of matched contrast is attributable to the cancellation of the topic distance component in the activation difference, which reduces the magnitude of the extracted direction below the threshold at which weight-matrix projection achieves measurable perturbation of the residual stream.

\item SVD orthogonalization of matched-contrast directions introduces an additional failure mode by distributing the already insufficient shared refusal signal across multiple orthogonal axes.
\end{enumerate}

The construction of contrast baselines for multi-directional refusal abliteration warrants further investigation.

\bibliography{references}

\begin{thebibliography}{10}
\providecommand{\natexlab}[1]{#1}
\providecommand{\url}[1]{\texttt{#1}}
\expandafter\ifx\csname urlstyle\endcsname\relax
  \providecommand{\doi}[1]{doi: #1}\else
  \providecommand{\doi}{doi: \begingroup \urlstyle{rm}\Url}\fi

\bibitem[Akiba et~al.(2019)Akiba, Sano, Yanase, Ohta, and
  Koyama]{akiba2019optuna}
Takuya Akiba, Shotaro Sano, Toshihiko Yanase, Takeru Ohta, and Masanori Koyama.
\newblock Optuna: A next-generation hyperparameter optimization framework.
\newblock In \emph{Proceedings of the 25th ACM SIGKDD International Conference
  on Knowledge Discovery and Data Mining}, pages 2623--2631, 2019.

\bibitem[Arditi et~al.(2024)Arditi, Obeso, Syed, Paleka, Panickssery, Gurnee,
  and Nanda]{arditi2024refusal}
Andy Arditi, Oscar Obeso, Aaquib Syed, Daniel Paleka, Nina Panickssery, Wes
  Gurnee, and Neel Nanda.
\newblock Refusal in language models is mediated by a single direction.
\newblock In \emph{Advances in Neural Information Processing Systems},
  volume~37, 2024.

\bibitem[Belrose et~al.(2023)Belrose, Schneider-Joseph, Ravfogel, Cotterell,
  Raff, and Biderman]{belrose2023leace}
Nora Belrose, David Schneider-Joseph, Shauli Ravfogel, Ryan Cotterell, Edward
  Raff, and Stella Biderman.
\newblock {LEACE}: Perfect linear concept erasure in closed form.
\newblock In \emph{Advances in Neural Information Processing Systems},
  volume~36, 2023.

\bibitem[{FailSpy}(2024)]{failspy2024abliterator}
{FailSpy}.
\newblock abliterator: Abliteration of refusal behavior from language models.
\newblock \url{https://github.com/FailSpy/abliterator}, 2024.

\bibitem[{kabachuha}(2025)]{kabachuha2025som}
{kabachuha}.
\newblock {SOM} integration for the {Heretic} abliteration framework.
\newblock GitHub Pull Request \#196, p-e-w/heretic, 2025.

\bibitem[{mlabonne}(2024)]{mlabonne2024harmless}
{mlabonne}.
\newblock harmless\_alpaca dataset.
\newblock Hugging Face: mlabonne/harmless\_alpaca, 2024.

\bibitem[Piras et~al.(2026)Piras, Mura, Brau, Oneto, Roli, and
  Biggio]{piras2025som}
Giorgio Piras, Raffaele Mura, Fabio Brau, Luca Oneto, Fabio Roli, and
  Battista Biggio.
\newblock {SOM} directions are better than one: Multi-directional refusal
  suppression in language models.
\newblock In \emph{Proceedings of the AAAI Conference on Artificial
  Intelligence}, volume~40, pages 32728--32736, 2026.

\bibitem[{Qwen Team}(2026)]{qwen2026qwen35}
{Qwen Team}.
\newblock Qwen 3.5.
\newblock \url{https://qwen.ai/blog?id=qwen3.5}, 2026.
\newblock Alibaba Cloud.

\bibitem[Weidmann(2024)]{weidmann2024heretic}
Philipp~Emanuel Weidmann.
\newblock Heretic: Automated {LLM} abliteration framework.
\newblock \url{https://github.com/p-e-w/heretic}, 2024.

\bibitem[Wollschl{\"a}ger et~al.(2025)Wollschl{\"a}ger, Elstner, Geisler,
  Cohen-Addad, G{\"u}nnemann, and Gasteiger]{wollschlager2025gradient}
Tom Wollschl{\"a}ger, Jannes Elstner, Simon Geisler, Vincent Cohen-Addad,
  Stephan G{\"u}nnemann, and Johannes Gasteiger.
\newblock The geometry of refusal in large language models: Concept cones and
  representational independence.
\newblock In \emph{Proceedings of the 42nd International Conference on Machine
  Learning}, 2025.

\end{thebibliography}

\end{document}